%% file: ms.tex
\documentclass[letterpaper]{article} 
\usepackage{aaai25}  
\usepackage{times}  
\usepackage{helvet}  
\usepackage{courier}  
\usepackage[hyphens]{url}  
\usepackage{graphicx} 
\usepackage{natbib}  
\usepackage{caption} 
\usepackage{algorithm}
\usepackage{algorithmic}

\usepackage{cleveref}
\usepackage{import}
\usepackage{subcaption}
\usepackage{multirow}
\usepackage{tabularx}
\usepackage{booktabs}
\usepackage{makecell}
\usepackage{twemojis}
\usepackage{pythonhighlight}
\usepackage{soul}
\usepackage{longtable}
\usepackage[inline]{enumitem}
\usepackage{xcolor}

\usepackage{newfloat}
\usepackage{listings}
\DeclareCaptionStyle{ruled}{labelfont=normalfont,labelsep=colon,strut=off} 
\floatstyle{ruled}
\newfloat{listing}{tb}{lst}{}
\floatname{listing}{Listing}
\title{COLUMBUS: Evaluating \underline{CO}gnitive \underline{L}ateral \underline{U}nderstanding through \underline{M}ultiple-choice re\underline{BUS}es}
\author {
    Koen Kraaijveld\textsuperscript{\rm 1},
    Yifan Jiang\textsuperscript{\rm 2},
    Kaixin Ma\textsuperscript{\rm 3},
    Filip Ilievski\textsuperscript{\rm 1}
}
\affiliations {
    \textsuperscript{\rm 1}Department of Computer Science, Faculty of Science, Vrije Universiteit Amsterdam\\
    \textsuperscript{\rm 2}Information Sciences Institute, University of Southern California\\
    \textsuperscript{\rm 3}Tencent AI Lab, Bellevue, WA\\
    h.j.kraaijveld@student.vu.nl, yjiang44@usc.edu, kaixinma@global.tencent.com, f.ilievski@vu.nl
}

\begin{document}

\maketitle

\begin{abstract}
While visual question-answering (VQA) benchmarks have catalyzed the development of reasoning techniques, they have focused on vertical thinking. Effective problem-solving also necessitates lateral thinking, which remains understudied in AI and has not been used to test visual perception systems. To bridge this gap, we formulate \textit{visual lateral thinking} as a multiple-choice question-answering task and describe a three-step taxonomy-driven methodology for instantiating task examples. Then, we develop \textsc{COLUMBUS}, a synthetic benchmark that applies the task pipeline to create QA sets with text and icon rebus puzzles based on publicly available collections of compounds and common phrases. \textsc{COLUMBUS} comprises over 1,000 puzzles, each with four answer candidates. While the SotA vision-language models (VLMs) achieve decent performance, our evaluation demonstrates a substantial gap between humans and models. VLMs benefit from human-curated descriptions but struggle to self-generate such representations at the right level of abstraction.

\end{abstract}

\begin{links}
    \link{Code}{https://github.com/koen-47/COLUMBUS}
\end{links}

\section{Introduction}

\begin{figure}
    \centering
    \includegraphics[width=0.75\linewidth]{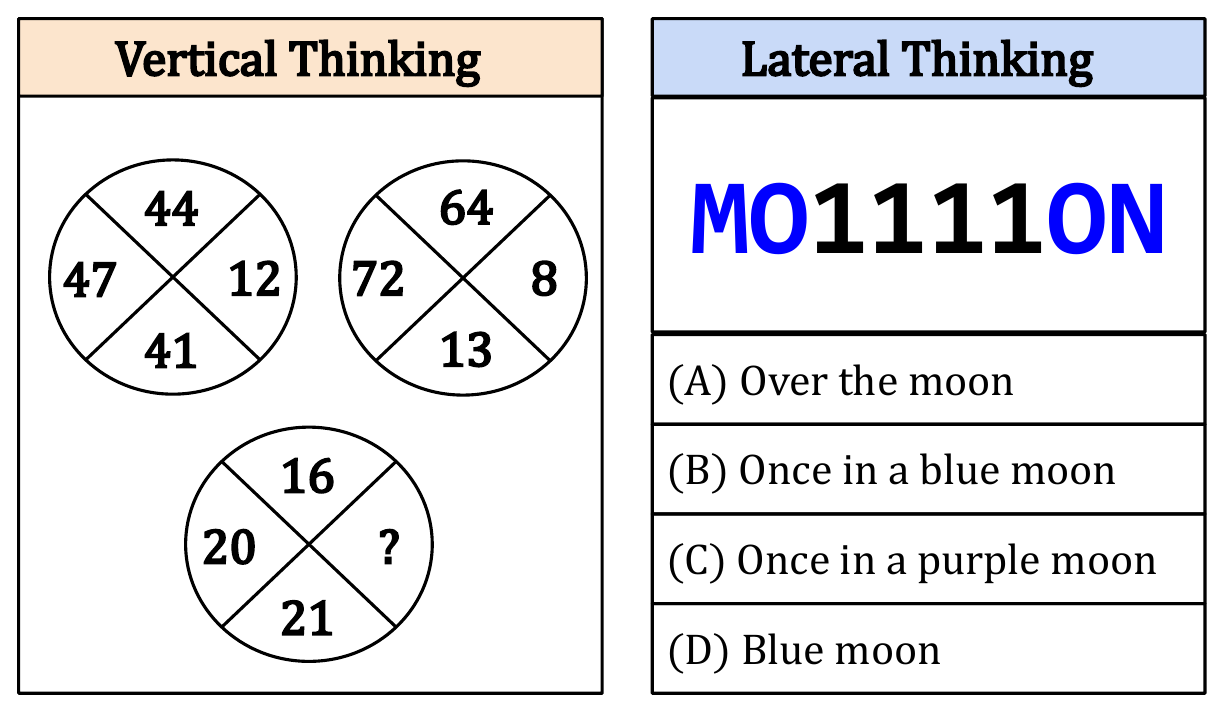}
    \caption{Left: vertical thinking puzzle from Machine Number Sense \cite{zhang_2020}. Right: lateral thinking rebus puzzle from our \textsc{COLUMBUS} benchmark.}
    \label{fig:vertical_lateral_puzzles_examples}
    \vspace{-1em}
\end{figure}

Human problem-solving seamlessly combines vertical and lateral thinking
\cite{de_bono_lateral_2016}. \textit{Vertical} thinking is an analytical search process that rewards logic, rules, and rationality. It optimizes correctness by narrowing down on quality solutions and rejecting suboptimal ones \cite{hernandez_vertical_2008}. For example, resolving the question mark in Figure \ref{fig:vertical_lateral_puzzles_examples} (left) requires systematically identifying that all examples adhere to the formula: (\textit{left} - \textit{top}) $\times$ \textit{right} + \textit{bottom} = 77.
Meanwhile, \textit{lateral} thinking \cite{de_bono_use_1971} is explorative, divergent, and creative \cite{hernandez_vertical_2008}. It expands the solution space by diverging into novel directions. As illustrated in the right part of Figure \ref{fig:vertical_lateral_puzzles_examples}, visual, spatial, verbal, and numerical cues must be interpreted unconventionally (defying common sense;~\citealp{jiang2024semeval}), a process that lends itself to lateral thinking. In this example of a \textit{rebus} puzzle, the numbers ``1111'' phonetically represent the word ``ONCE'', while the visual-spatial relationship between the blue letters ``MO'' and ``ON'' spell ``BLUE MOON''. As ``ONCE'' is placed inside ``BLUE MOON'' this leads to the solution \textit{B) Once in a blue moon}.

Existing benchmarks for visual question answering (VQA) \cite{agrawal_2016} have been instrumental in exploring and enhancing the vertical thinking skills of vision-language models (VLMs). Popular subtasks are visual reasoning \cite{johnson_clevr_2017, li_qlevr_2022, li_super_2023}, abstract visual reasoning (AVR) \cite{chollet_2019, malinowski_2015, malkinski_2023, zhang_2019}, and visual commonsense reasoning (VCR) \cite{bitton-guetta_breaking_2023}, all requiring both processing of visual as well as linguistic information. Meanwhile, lateral thinking benchmarks have recently been proposed as word and sentence puzzles but are limited only to the textual modality \cite{jiang_brainteaser_2023, huang_lateval_2023}. Hence, there is a lack of lateral thinking benchmarks for multimodal settings combining text and vision.

To this end, we study \textit{how well VLMs exhibit multimodal lateral thinking}. Our contributions are as follows:

\begin{enumerate}
    \item \textbf{A taxonomy-driven three-step methodology} for creating lateral thinking tasks in a multiple-choice VQA format. Our taxonomy definition yields 18 rules that manipulate the visual attributes and relationships of the puzzle's elements (text or icons). The puzzle rendering step leverages this taxonomy to create a graph representation for a puzzle answer and generate an image for the graph. The distractor sampling step is based on a weighted average of orthographic and semantic similarity between a puzzle's correct answer and its visible elements.
    
    \item \textbf{A synthetic benchmark called \textsc{COLUMBUS}} that applies the lateral thinking methodology to create QA sets with rebus puzzles based on public collections of phrases (e.g., idioms) and compound words.\footnote{The name refers to the demonstration of lateral thinking in the story of \textit{Columbus' Egg} \cite{benzoni_history_2017}.} \textsc{COLUMBUS} comprises over 1,000 puzzles consisting of textual and icon elements, each with four answer candidates. 
    
    \item \textbf{An experimental analysis} with \textsc{COLUMBUS} with human participants and representative state-of-the-art (SotA) vision-language models evaluated in a zero-shot setting, revealing that models perform decently but lag behind humans. Moreover, models benefit from human-curated descriptions, but even the SotA ones struggle to generate representations at the right level of abstraction.
\end{enumerate}

\section{Related Work}

\noindent \textbf{Rebus Puzzles.}
In psychology, rebus puzzles have been known to demand lateral thinking  \cite{salvio_2016, threadgold_2018, macgregor_2008}. Prior work \cite{salvio_2016,threadgold_2018} reports human accuracies of 74.5\% and 53.31\%, respectively, and compares the impact of vertical and lateral thinking, concluding that using lateral thinking led to a significant improvement in the number of puzzles solved. To our knowledge, the only existing benchmark of rebus puzzles that assesses VLMs is REBUS \cite{gritsevskiy_rebus_2024}. This benchmark contains 333 human-annotated puzzles separated into 13 categories with three difficulty levels. Half of the models tested in this work achieve less than 5\% accuracy. The authors ascribe this difficulty to the benchmark's reliance on world knowledge (e.g., cities, towns, public transport stations) and vertical thinking skills like string manipulation. Instead, we devise a methodology for automatic and scalable generation of rebus puzzles based on publicly available resources. The sources to create \textsc{COLUMBUS} (phrases and compounds) are deliberately selected to focus on lateral thinking only and minimize the need for world knowledge and vertical thinking.

\noindent \textbf{Vertical Thinking in VQA.}
AVR puzzles, illustrated in Figure \ref{fig:vertical_lateral_puzzles_examples} (left), are commonly used to assess multimodal reasoning. Discriminative tasks such as Raven's Progressive Matrices \cite{raven_1941, barrett_2018, zhang_2019} and Visual Analogy Problems \cite{webb_learning_2020} involve completing sequences of panels with abstract shapes selected from a predefined set of options. Bongard problems \cite{bongard_recognition_1968, nie_bongard-logo_2020} require discovering the rules that separate and govern shapes across two sets of panels, though these rules must be described in natural language. MARVEL \cite{jiang_marvel_2024} encompasses these benchmarks with a more comprehensive set of patterns, input shapes, and configurations, along with rigorous checks to assess that model answers are grounded in perception and reasoning. Alternatively, generative approaches, like the Abstraction and Reasoning Corpus \cite{chollet_2019}, test the ability to recreate missing panels without choosing from predefined options. 
A comprehensive review of AVR puzzles is provided by \citeauthor{malkinski_2023}~(\citeyear{malkinski_2023}). Rather than using puzzles, CLEVR \cite{johnson_clevr_2017}, QLEVR \cite{li_qlevr_2022}, and Super-CLEVR \cite{li_super_2023} are synthetic benchmarks that test logical reasoning by analyzing 3D rendered scenes of objects. WHOOPS! \cite{bitton-guetta_breaking_2023} is a visual commonsense reasoning benchmark of images generated through diffusion models that consist of illogical scenarios (e.g., Albert Einstein holding a smartphone). Crucially, these benchmarks rely on vertical thinking and do not test out-of-the-box thinking. Thus, our lateral task methodology and the \textsc{COLUMBUS} benchmark enable a complementary assessment of the models' abilities.

\noindent \textbf{Text-based Lateral Evaluation.}
Recent work has recognized an analogous omission of lateral thinking for the text domain. \citeauthor{jiang_brainteaser_2023}~(\citeyear{jiang_brainteaser_2023}) introduce \textsc{BrainTeaser}, a multiple-choice lateral thinking benchmark with 1,100 puzzles adapted from online sources. \textsc{BrainTeaser} requires models to bypass commonsense associations to arrive at the correct answer. Similarly, \citeauthor{huang_lateval_2023}~(\citeyear{huang_lateval_2023}) present LatEval, a benchmark consisting of 300 lateral puzzles. Each puzzle in LatEval is an interactive game between two large language models (LLMs) in which the LLM under evaluation must solve the puzzle presented by the host LLM. While we share the goal of testing models' lateral thinking ability, we broaden the evaluation scope to a multimodal setting covering text and vision.

\section{Methodology for Visual Lateral Tasks}
\label{section_methodology}

\begin{figure}[!b]
    \centering
    \includegraphics[scale=0.4]{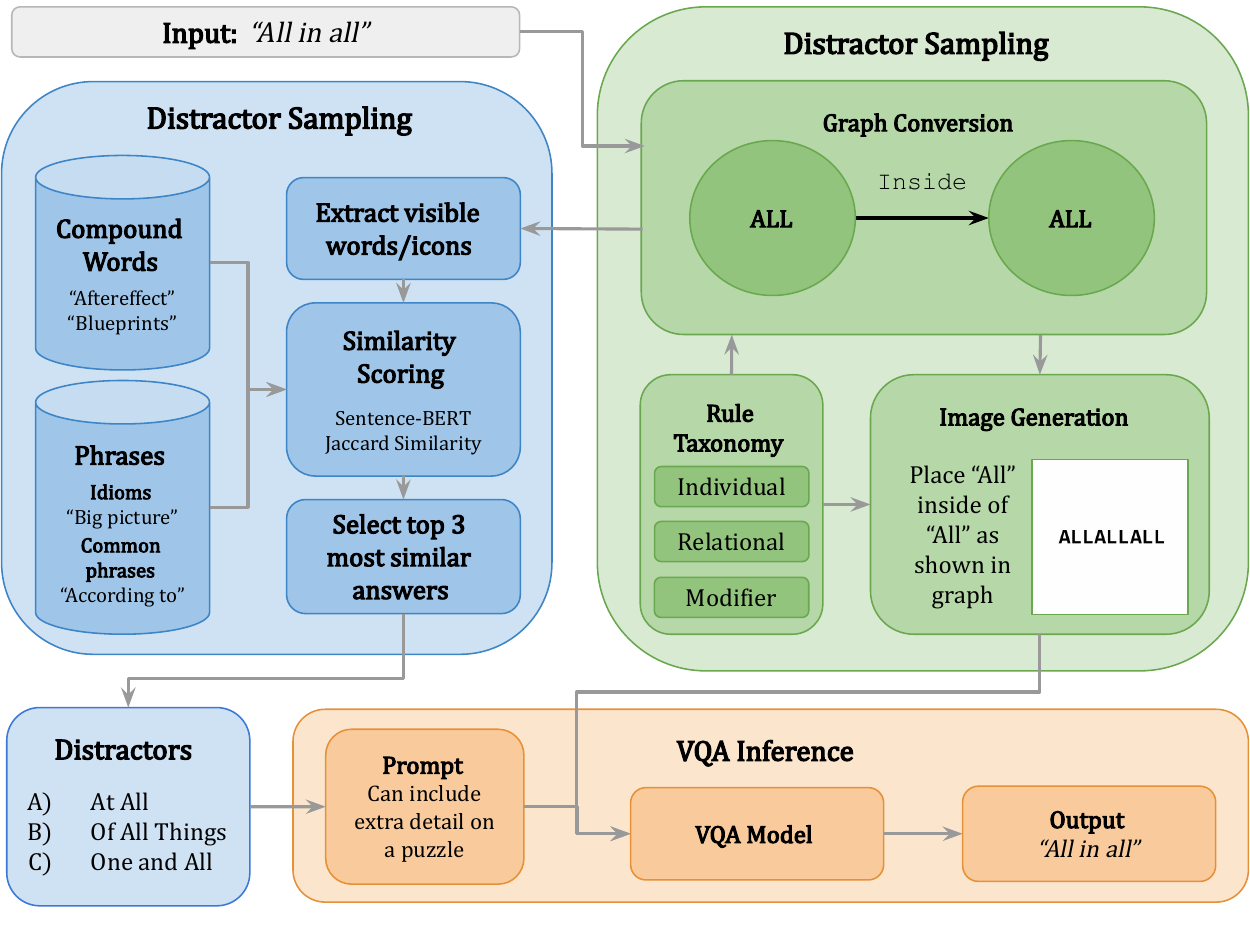}
    \caption{Methodology for visual lateral thinking tasks.}
    \label{fig:pipeline_overview}
\end{figure}

To ensure a straightforward automatic evaluation and minimize answer ambiguity, we frame each puzzle as a multiple-choice VQA pair. A puzzle $p = (I, (q, O, c))$ consists of a rebus image $I \subseteq \mathcal{I}$ and question $q \in \mathcal{S}$ with correct answer $c \in \mathcal{S}$ chosen from options $O = \{o_1, o_2, o_3, c\}$; $O \subseteq \mathcal{S}$. $\mathcal{I}$ and $\mathcal{S}$ denote the space of images and strings, respectively. Each $I$ can be decomposed into a set of elements $E \subseteq \mathcal{I} \cup \mathcal{S}$, where $\forall e \in E$ $(e \in \mathcal{I} \oplus e \in \mathcal{S})$.
The latent rules that govern the appearance and visual-spatial relationships of each $e \in E$ are determined by $R : \mathcal{S} \rightarrow \mathcal{I}$. The goal in solving $p$ is to select a response $r \in O$ such that $R(r) = R(c)$.

\begin{figure*}[!ht]
    \centering
    \includegraphics[width=0.85\linewidth]{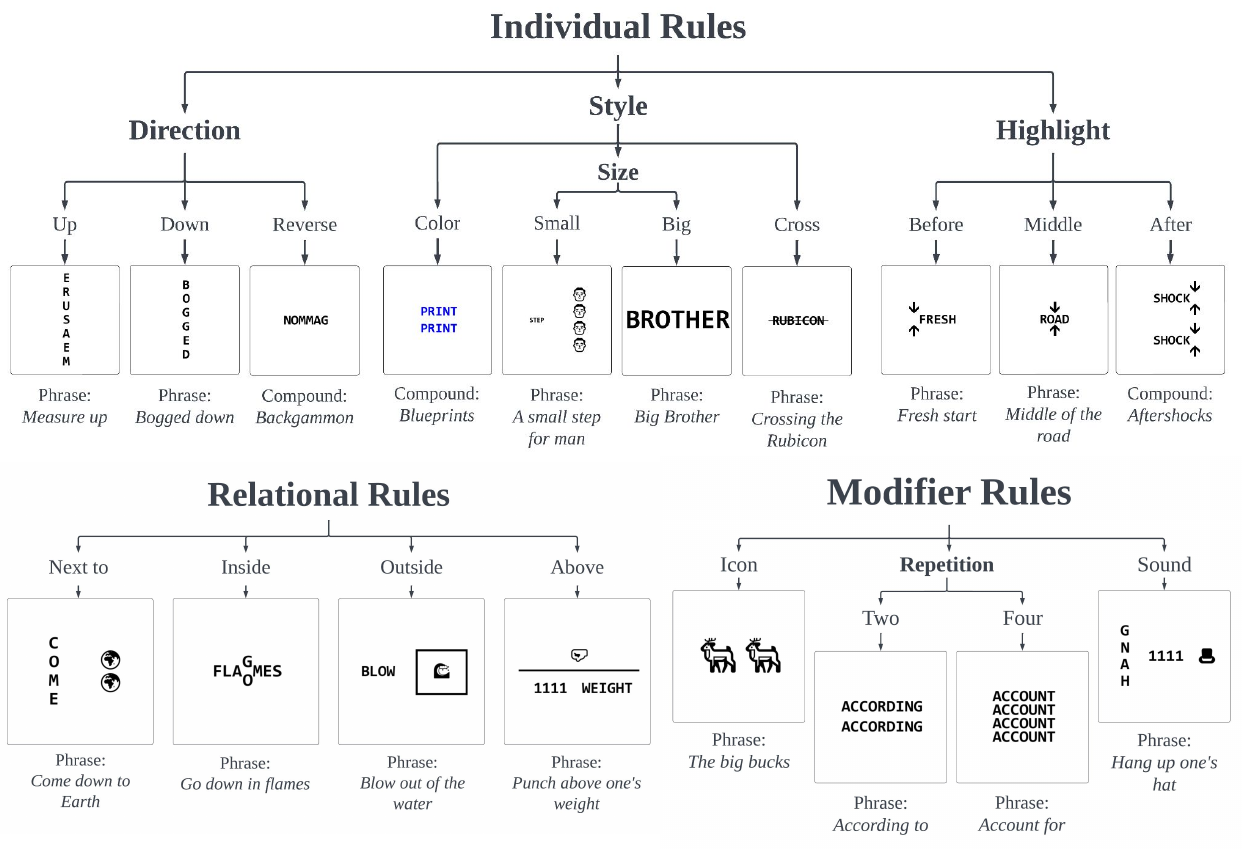}
    \caption[Three taxonomies]{Three taxonomies that classify and organize the \textit{individual} (top), \textit{relational} (bottom left), and \textit{modifier} (bottom right) rules used to manipulate the appearance and position of elements in a rebus puzzle. For each rule, we present an example puzzle and its answer, both taken directly from \textsc{COLUMBUS}.}
    \label{fig:rule_taxonomies}
\end{figure*}

Figure \ref{fig:pipeline_overview} depicts our method. 
As rebus puzzles are typically built around idiomatic expressions, compound words, or common phrases (e.g., ``according to'', ``a bit too much'') \cite{salvio_2016, threadgold_2018}, we assume phrases and compounds as inputs for our method.
We start by designing a taxonomy of latent rules. Using this taxonomy, each compound or phrase is converted into an attributed, directed graph, which is subsequently converted into the puzzle image. Finally, distractors are sampled by identifying other compounds or phrases that overlap with, or are semantically similar to, the method input.

\subsection{Taxonomy of Latent Rules} \label{section_taxonomy}
We derive a novel taxonomy of latent rules to support the development of lateral thinking puzzles. The taxonomy consolidates online guides and databases of rebus puzzles and a rebus categorization scheme outlined by \citeauthor{salvio_2016}~(\citeyear{salvio_2016}). We manually select and organize the categories in these sources such that each rule uniquely manipulates an element through visual, spatial, verbal, and numerical properties. We ensure that each rule can be automatically operationalized and mixed with others in the same puzzle.

The resulting taxonomy (\Cref{fig:rule_taxonomies}) consists of 18 rules, grouped into three categories according to how they manipulate elements in a puzzle:
\begin{enumerate*}
    \item \textbf{Individual} rules define the unary characteristics of an element in a rebus. Example rules include reversing character order (direction), the text color (style), and adding arrows before the element (highlight).  
    \item \textbf{Relational} rules define the positioning between a pair of elements. We define four \textit{relational} rules, placing an element beside/inside/above/outside another.
    \item \textbf{Modifier} rules are designed to be mutually inclusive with other \textit{individual} rules. Examples include repeating an element multiple times or substituting it with a phonetically similar element. 
\end{enumerate*}

\subsection{Puzzle Rendering} \label{section_puzzle_generation}
Rebus puzzles include elements (i.e., text or icons) whose appearance and position are determined by latent rules triggered by specific keywords in the puzzle's answer.
This is illustrated in Figure \ref{fig:vertical_lateral_puzzles_examples} (right), where the words ``ONCE'', ``IN'', ``BLUE'', and ``MOON'' determine the puzzle's elements and their arrangement. We expect that SotA generative models cannot be reliably applied to generate rebus puzzles, a hypothesis we validate in \Cref{section_rebus_generation_results}.
Instead, we render a puzzle by a taxonomy-driven transformation of its input elements, which first produces a graph and subsequently an image (green part in Figure \ref{fig:pipeline_overview}).

\begin{figure}[!t]
    \centering
    \begin{subfigure}[h]{0.5\textwidth}
        \centering
        \includegraphics[scale=0.225]{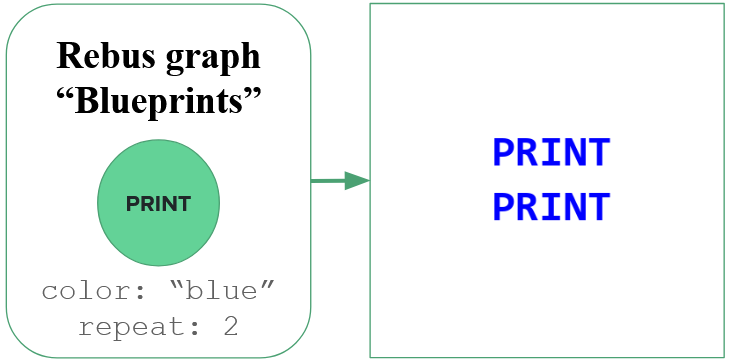}
        \label{fig:rebus_graph_example_1}
    \end{subfigure}
    \\ \vspace{0.2cm}
    \begin{subfigure}[h]{0.5\textwidth}
        \centering
        \includegraphics[scale=0.225]{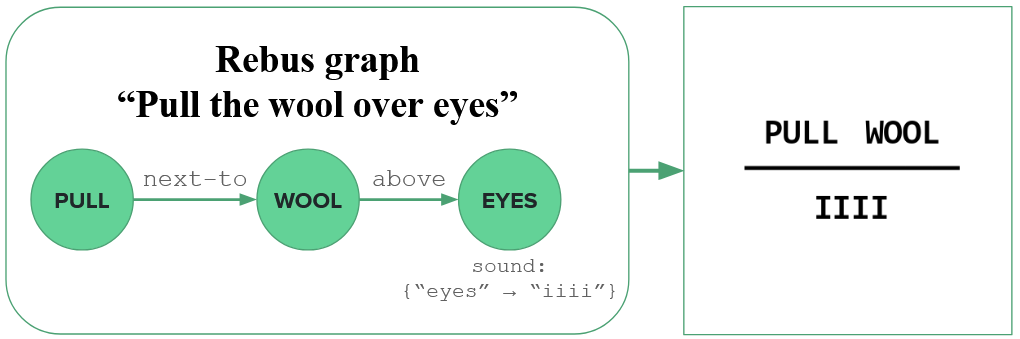 }
        \label{fig:rebus_graph_example_3}
    \end{subfigure}

    \caption{Two examples of directed attributed graphs (left) representing rebus puzzles (right).}
    \label{fig:rebus_graph_examples}
\end{figure}

\noindent \textbf{Graph Generation Algorithm.} We generate a directed, attributed graph whose nodes are elements that will be rendered into a puzzle image. The node attributes specify the rendering of that element, i.e., the \textit{individual} or \textit{modifier} rules that will apply to it. The edges between two nodes are annotated with an attribute that specifies their \textit{relational} rule. 
We parse a puzzle answer into a graph by following a separate procedure for compounds and phrases. \Cref{fig:rebus_graph_examples} shows the rebus graphs for two puzzle images based on a compound and a phrase, respectively. 
    
For \textit{compounds} (Figure \ref{fig:rebus_graph_examples} top), we create a graph with a single node using the following steps. First, we split a compound into its constituent words, e.g., ``blueprints" consists of ``blue'' and ``prints''. For each word, we check if it matches against any of the keywords of an \textit{individual} rule, e.g., ``blue'' triggers the \textit{color} rule.
Then, we create a graph with a single node consisting of the other constituent word (``prints'') and set this node's color attribute to blue. Since we detect that ``prints'' is plural, we set its \textit{repetition} modifier attribute to 2. For the final step, we check if the word in the node corresponds to any available homophones or icons, which does not occur in this case. In cases where both constituent words of a compound trigger a rule, we generate both graph interpretations of the input.

For \textit{phrases} (\Cref{fig:rebus_graph_examples} bottom), we first identify keywords belonging to a \textit{relational} rule, e.g., ``over'' triggers the \textit{above} rule. At this word, we split the phrase into two substrings: ``pull the wool over eyes'' yields ``pull the wool'' and ``eyes''. On each substring, we run the compound parser over each pair of words from left to right, e.g., the first and only pair of the first substring is (``pull'', ``wool''). This process results in a set of nodes, which are then connected using the \textit{next to} rule, yielding two path subgraphs. The final step involves connecting these two subgraphs with an edge with the \textit{relational} rule identified in the first step (\textit{above} in our example). 



\noindent \textbf{Image Generation.} We select one of four templates using only the graph as input. Templates include $x,y$ coordinates and a font-size multiplier. Three templates position up to three points equally along the $x$-axis in the image center, mapping graph nodes (left to right) to template points. The fourth handles graphs with the \textit{above} rule. We change the appearance of the elements in an image according to the attributes of that element's respective node.

\subsection{Distractor Sampling}
Distractor sampling (blue part in Figure \ref{fig:pipeline_overview}) selects the three most similar compounds or phrases to the input semi-automatically. We opt for sampling rather than data augmentation approaches like rephrasing because compounds and proverbial phrases are challenging to generate automatically. To select a distractor for a puzzle, we obtain its visible elements (text and icons) from the graph representation and compute similarity to all other phrases/compounds.
The similarity uses a $\lambda$-weighted combination of Jaccard word overlap \cite{leskovec_mingin_2014} and cosine similarity using Sentence-BERT embeddings \cite{reimer_sentence-bert_reimers}.
We expect that distractors with word overlaps make the task more challenging because the test-taker works with the visible words. 
Since word overlap may fail to select relevant distractors when the visible words only occur once or too many times across the entire dataset of phrases and compounds, we also leverage semantic cosine similarity to include distractors that contain synonyms of the words in the original input.

\section{The \textsc{COLUMBUS} Benchmark} \label{section_columbus_benchmark}

We apply our proposed pipeline to instantiate the first visual lateral thinking benchmark, \textsc{COLUMBUS}. 

\noindent \textbf{Puzzle Answer Collection.} We start by scraping common English phrases from publicly available sources, namely Wiktionary and \url{www.theidioms.com}, yielding 9,745 instances. We use the Large Database of English Compounds (LaDEC) \cite{gagne_ladec_2019} for compound words. This dataset has been feature-engineered and curated by humans, consisting of 8,957 compounds. We fill rules that appear less than ten times across the benchmark by semi-automatically adding compounds and phrases that trigger them, with the assistance of prompting the OpenAI's ChatGPT-3.5 model \cite{brown_language_2020}. All combined, we collect 18,836 candidate answers from which to generate puzzles. Additionally, we collected homophones (25 samples) and icons (480 samples). Homophones were added manually by recognizing common ones found in rebus puzzle databases. The icon collection combines icons scraped from an online source and manually added ones.\footnote{Source: \url{https://unicode.org/Public/emoji/11.0/emoji-test.txt}}

\noindent \textbf{Quality Control.} 
The graph parsing for all phrases and compounds includes a preprocessing step to remove stopwords that do not belong to the set of rule-triggering keywords.\footnote{E.g., ``to" is a stopword, but triggers the \textit{repetition: two} rule phonetically. 
} Multiple elements with \textit{individual} rules can still be present in the same puzzle, and more than one \textit{modifier} rule can be applied to an element. However, we apply at most one \textit{individual} rule to a single element. In cases where multiple \textit{individual} rules can be applied to a single element, we generate these individually for each rule as separate puzzles. To further improve readability and limit the risk of overlapping elements, we restrict the image's rendered elements using heuristics based on the number of elements and their rules. These exclude images generated on graphs that (1) have more than three nodes connected by the \textit{next to} rule, (2) have an \textit{above} rule where the top/bottom exceeds two nodes, (3) have an \textit{above} rule where either the top/bottom exceeds two nodes, (4) have an \textit{outside} rule where either the inside/outside portion has more than one node. We take the top 1,000 from the remaining puzzles with the most edges and rules per node to ensure the benchmark is challenging. 
To provide a fairer comparison between textual and icon puzzles, each puzzle containing an icon is duplicated, and all its icons are converted to their textual counterparts. Finally, we filter out low-quality puzzles with overlapping or overflowing text from this remaining set.

\noindent \textbf{Benchmark Composition.}
We split the benchmark into two partitions: \textsc{COLUMBUS-text} that only contain text and \textsc{COLUMBUS-icon} that contain at least one icon. Between these two partitions, \textsc{COLUMBUS} features an \textit{overlap} subset of 338 puzzle pairs. Each pair consists of two versions of the same puzzle: one version uses icons, and the other uses text instead of those icons. Table \ref{tab:puzzle_statistics} shows key statistics about \textsc{COLUMBUS}.  While non-icon puzzles are more numerous, icon puzzles feature more elements. This can be attributed to the difference in the answer length, as longer answers feature more chances that a word can be replaced with an icon. 

\begin{table}[!ht]
    \small
    \centering
    \caption{Key statistics of the \textsc{COLUMBUS} benchmark.}
    \setlength{\tabcolsep}{1mm}
    \begin{tabular}{@{}llrrr@{}}
        \toprule
        Category & Statistic & \makecell[l]{\textsc{text}} & \makecell[l]{\textsc{icon}} & All \\
        \midrule
        \multirow{2}{*}{General} & Number of puzzles & 634 & 371 & 1,005 \\
        
        & \makecell[l]{Mean answer length (\# words)} & 4.12 & 5.35 & 4.58 \\
        \midrule
        
        \multirow{3}{*}{Rules} & Freq. of \textit{individual} rules & 253 & 76 & 329 \\
        & Freq. of \textit{relational} rules & 540 & 425 & 965 \\
        & Freq. of \textit{modifier} rules & 503 & 255 & 758 \\
        \midrule
        
        \multirow{3}{*}{Graphs} & Freq. of single node graphs & 197 & 35 & 232 \\
        & Freq. of double node graphs & 332 & 246 & 578 \\
        & Freq. of triple node graphs & 105 & 90 & 195 \\
        \midrule
        
        Distractors & \makecell[l]{\% of questions with distractors\\ containing visible puzzle words} & 89.27 & 97.57 & 92.34 \\
        \bottomrule
    \end{tabular}
    \label{tab:puzzle_statistics}
\end{table}

\section{Experimental Design}

\subsubsection{Model Families.} \label{section_model_selection} 
We include open- and closed-source instruction-tuned and non-instruction-tuned VLMs. We also test closed-source models enriched with forward and backward chaining.
We evaluate all models in a zero-shot setting using standard hyperparameter values.

For non-instruction-tuned models, we test 1) \textit{BLIP-2} \cite{li_blip-2_2023} with the OPT-2.7b and the OPT-6.7b LLMs \cite{zhang_opt_2022}; 2) \textit{Fuyu-8b} \cite{bavishi_fuyu-8b_2023}, a multimodal text and image transformer that achieves competitive performance on VQA tasks. We also evaluate \textit{CLIP} \cite{radford_learning_2021}, a seminal VLM and a foundation for many other models used in our experiments. As CLIP is not a VQA model, we switch its task to image classification, which must match the image of a puzzle to the correct answer from the four available choices.

As instruction-tuned models, we include 1) BLIP-2 coupled with \textit{Flan-T5-11b} \cite{chung_scaling_2022}, which achieves SotA performance on zero-shot VQA tasks; 2) \textit{InstructBLIP} \cite{dai_instructblip_2023}, an instruction tuned version of BLIP-2 model that uses Vicuna-7b \cite{zheng_judging_2023}; 3) \textit{QwenVL} \cite{bai_qwen-vl_2023}, a 7 billion visual multimodal version of the Qwen LLM \cite{bai_qwen_2023} from which we use the chat variant; 4) \textit{CogVLM} \cite{wang_cogvlm_2023}, a 17 billion parameter VLM that achieves SotA performance on several VQA benchmarks; 5) \textit{Llava} \cite{liu_visual_2023}, a large VLM that achieves SotA performance on several vision benchmarks despite its lack of billion-scale data. For Llava, we use the 13b (v1.5) and 34b (v1.6) variants; 6) \textit{Mistral-7b} (v2) \cite{jiang_mistral_2023} to use in text-only, question-answering (QA) auxiliary experiments.

For closed-source models, we selected four models from two representative families based on their promising performance in public visual reasoning benchmarks~\cite{lu2023mathvista,liu2023mmbench}: 1)~GPT-4o and GPT-4o-mini~\cite{gpt4_2023} and 2)~Gemini 1.5~(Pro) and Gemini 1.5~(Flash)~\cite{geminiteam_2023}.

We experiment with two \textit{structural} variants of closed-source models: forward and backward chaining~\cite{UBMA_285413791}.
In \textit{forward chaining (FC)}, the model constructs evidence from an image and connects this evidence with the optimal candidate answer~\cite{wang2024candle}.
The forward chaining approach first prompts models to generate JSON files with attributes (e.g., name, relation, description) for each visible puzzle element, which can later be used as a reference in the final prompting.
As a representative of \textit{backward chaining (BC)} from question and image towards the answer, we test \textit{belief graphs} \cite{kassner_2023} where a model derives and evaluates explanations for each candidate answer.
Belief graphs excavate additional information by recursively evaluating the truth assignments of premises generated for each answer candidate. The truth assignments are then optimized with an SAT solver, yielding the most probable answer. This approach is evaluated on a random subset of 50 puzzles.

\noindent \textbf{Human Evaluation.} To estimate human performance on \textsc{COLUMBUS}, we ask five participants to answer a subset of 103 randomly selected puzzles, consisting of 37 text, 40 icon puzzles, and 13 \textit{overlap} puzzles with both a textual and icon variant. 

\noindent \textbf{Model Inputs.}
We explore four human-curated input levels, each providing the model with increasing information about the puzzle, i.e., its description and details on the nodes or edges of a puzzle's graph. 
Specifically:
\begin{enumerate*}
    \item no description of the nature of the puzzle, nor the graph;
    \item only a description of the nature of the puzzle;
    \item description of the nature of the puzzle and the graph nodes;
    \item description of the nature of the puzzle and the full graph (both nodes and edges).
\end{enumerate*}

\noindent \textbf{Evaluation Protocol.}
Following other multiple-choice benchmarks \cite{jiang_brainteaser_2023, zhu_visual7w_2016, de_faria_2023}, we use accuracy as the evaluation metric, defined as the percentage of puzzles solved correctly. 
To extract answers from a model's output, we use regex to check for choice symbols (e.g., ``A'') if they are present and then perform exact string matching to the correct answer/symbol. For the larger, more flexible models that produce long explanations for their answers, we use GPT-4o to extract their answers automatically. For model outputs that answer a given puzzle with multiple options, we pick one of them randomly. All models are run three times, and their performance is averaged to account for randomness.

\section{Results}

We investigate five questions: 1) How well can VLMs solve rebus puzzles that require lateral thinking? 2) Can forward and backward chaining enhance lateral thinking performance? 3) Do VLMs benefit from prompts that supply more information about the puzzle?
4) How does the performance of VLMs vary across different puzzle rules?
5) Can VLMs generate task puzzles directly?

\begin{table}[!ht]
    \small
    \centering
    \caption{Results for each model on \textsc{COLUMBUS-text} and \textsc{COLUMBUS-icon}. The accuracy's mean and standard deviation (SD) are reported across three runs. The highest and second highest model results are highlighted in \textbf{bold} and \underline{underlined}, respectively. The prompt includes a description of the nature of the puzzle (i.e., prompt 2). 
    We did not test backward chaining for the Gemini models, as they do not output probabilities.}
    \begin{tabular}{@{}lrrrr@{}}
        \toprule
        \multirow{2}{*}{{\textbf{Model}}} & \multicolumn{2}{c@{}}{\makecell[c]{\textsc{text}}} & \multicolumn{2}{c@{}}{\makecell[c]{\textsc{icon}}} \\
        \cmidrule(lr){2-3} \cmidrule(lr){4-5}
        
        & \makecell[l]{Mean} & \makecell[l]{SD} & \makecell[l]{Mean} & \makecell[l]{SD} \\
        \midrule
        
        \makecell[l]{CLIP} & 56.15 & 0.00 & 52.56 & 0.00 \\
        
        \makecell[l]{BLIP-2 OPT (2.7b)} 
        & 24.74 & 0.21 & 24.08 & 0.13 \\
        
        \makecell[l]{BLIP-2 OPT (6.7b)} 
        & 23.95 & 0.16 & 25.61 & 0.00 \\
        
        Fuyu (8b) & 32.02 & 0.00 & 31.00 & 0.00 \\
        \midrule
        
        \makecell[l]{InstructBLIP Vicuna (7b)} 
        & 51.47 & 0.13 & 51.75 & 0.38 \\
        
        Qwen-VL (7b) 
        & 58.02 & 0.35 & 63.16 & 0.55 \\    
        
        \makecell[l]{BLIP-2 Flan-T5-XXL (11b)} 
        & 68.24 & 0.07 & 71.97 & 0.00 \\
        
        Llava (13b) 
        & 58.02 & 0.09 & 58.76 & 0.00 \\
        
        CogVLM (17b) 
        & 59.28 & 0.09 & 60.11 & 0.00 \\
        
        Llava (34b) 
        & 66.82 & 0.73 & 73.13 & 1.41 \\
        \midrule
        
        GPT-4o & \underline{80.89} & 0.97 & \textbf{83.34} & 1.11 \\
        GPT-4o-mini & 73.96 & 0.71 & 77.69 & 0.49 \\
        
        \makecell[l]{Gemini 1.5 (Pro)} & 71.56 & \underline{3.71} & 77.52 & \textbf{5.08} \\
        \makecell[l]{Gemini 1.5 (Flash)} & 64.42 & 2.00 & 67.44 & 2.93 \\
        
        \midrule
        GPT-4o (FC) & \textbf{81.28} & 1.15 & \underline{79.20} & 0.78 \\
        GPT-4o-mini (FC) & 73.53 & 0.79 & 74.36 & 1.29 \\
        \makecell[l]{Gemini 1.5 (Pro) (FC)} & 69.98 & 3.00 & 72.10 & 3.17 \\
        \makecell[l]{Gemini 1.5 (Flash) (FC)} & 72.00 & 1.42 & 75.88 & 3.55 \\
        
        GPT-4o (BC) & 64.37 & 1.63 & 71.67 & \underline{4.71} \\
        GPT-4o-mini (BC) & 45.93 & \textbf{8.65} & 60.00 & 4.08 \\ \midrule
        
        Human & 98.00 & N/A & 93.21 & N/A \\
        \bottomrule
    \end{tabular}
    \label{tab:results_main_table}
\end{table}

\subsection{Overall Performance} 
Table \ref{tab:results_main_table} shows the performance of each model on \textsc{COLUMBUS-text} and \textsc{COLUMBUS-icon}. The closed-source and larger open-source models perform best on both partitions, while the small, non-instruction-tuned models perform near-randomly. 
Comparing the mean model performance with text and icons, we see no significant difference. Namely, the accuracy is slightly higher on \textsc{COLUMBUS-icon}, whereas on the overlapping set of 338 puzzles, we observe a slightly higher accuracy on the textual puzzles. As expected, the best model for each partition is consistently GPT-4o. 
Yet, none of the models surpass human accuracy, with average gaps of 38.17\% on \textsc{COLUMBUS-text} and 30.64\% on \textsc{COLUMBUS-icon}.

\begin{figure}[!t]
    \centering
    \includegraphics[scale=0.27]{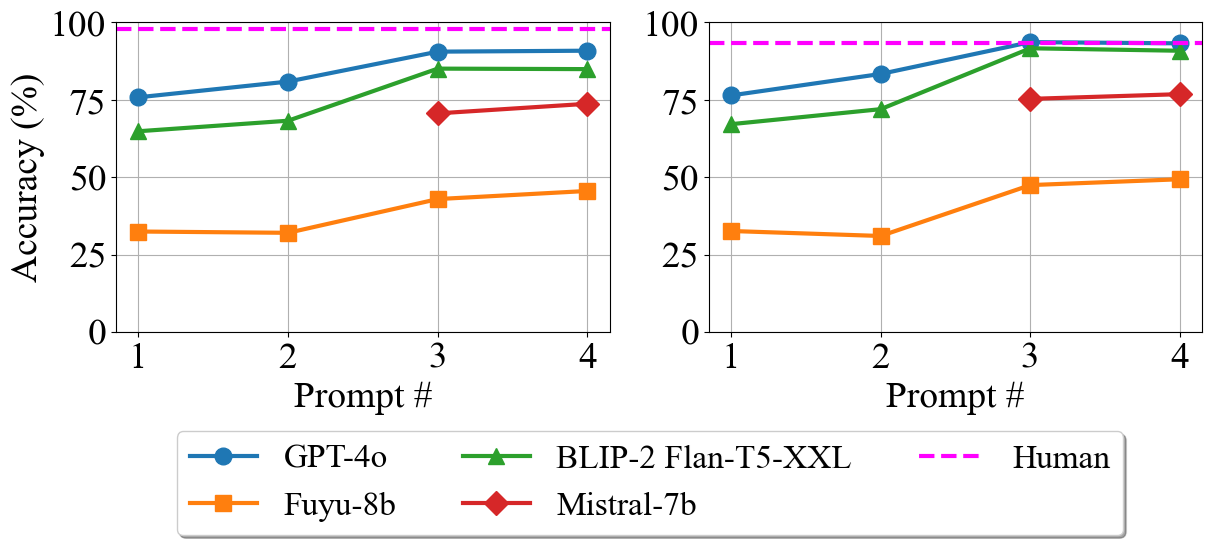}
    \caption{Results for four prompts that supply the model with increasing information for \textsc{COLUMBUS-text} (left) and \textsc{COLUMBUS-icon} (right) (averaged across three runs). The best-performing model from the following types is shown: open-source non-instruction VLM (Fuyu-8b), open-source instruction VLM (BLIP-2 Flan-T5-XXL), closed-source VLM (GPT-4o), and text-only LLM (Mistral-7b), as well as human accuracy.}
    \label{fig:results_all_prompts_four_models}
\end{figure}

\subsection{Impact of Structural Reasoning}

The two structured variants show opposing results (Table \ref{tab:results_main_table}). Forward chaining,
leading the model to generate graph descriptions in JSON format, yields little effect on the performance of GPT-4o (-1.88\%) and Gemini (+2.26\%), averaged across both models and partitions. Both models suffer from a gap against human performance. On the contrary, backward chaining yields an 14.1\% and 22.86\% drop in accuracy for GPT-4o and GPT-4o-mini averaged across the two partitions. We ascribe this to the models lacking a global overview of the image, as each evaluated premise focuses on local parts of a puzzle without cohesively pointing towards a candidate answer.


\subsection{Model Sensitivity to Input Information}
Can models benefit from a ground-truth structured description of the puzzle provided in their input?
Figure \ref{fig:results_all_prompts_four_models} shows that adding information about the nature of the puzzle (prompt 2) has little effect (+2.68\% and +3.39\% for textual and icon puzzles, respectively). Adding a description of the graph nodes (prompt 3) increases the model performance by $11.91\%$ and $14.9\%$ for non-icon and icon puzzles, respectively, reaching over 90\% for GPT-4o. However, adding information on the \textit{relational} rules only increases performance $1.47\%$ and $0.56\%$ for COLUMBUS-\textsc{text} and -\textsc{icon}, respectively. 
Considering the example in Figure \ref{fig:vertical_lateral_puzzles_examples}, the models extract the text as is (e.g., extract ``MO111ON'') and cannot make the lateral connection that words/letters need rearranging. Even GPT-4o struggles consistently with this, such as with certain \textit{direction} rules, as discussed in the next Section. 

\begin{figure*}[!ht]
    \centering
    \includegraphics[width=0.91\linewidth]{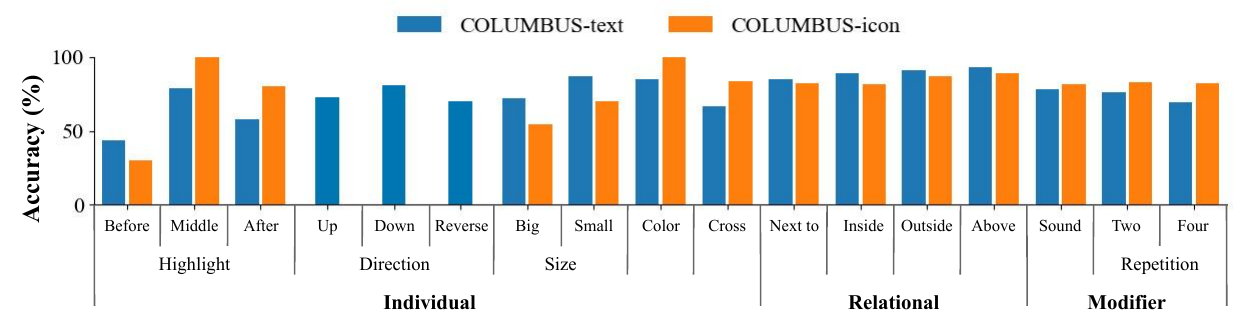}
    \caption{Percentage of puzzles solved by GPT-4o for a single run in COLUMBUS-\textsc{text} and COLUMBUS-\textsc{icon} for each rule. The prompt used only describes the nature of the puzzle~(prompt 2). For COLUMBUS-\textsc{icon}, the \textit{direction} rules are omitted because these either do not function with icons or become functionally identical to other rules when combined with icons.}
    \label{fig:results_rule_analysis_icons_no_icons}
\end{figure*}

\subsection{Rule-based Analysis} \label{section_rule_based_analysis}
Figure \ref{fig:results_rule_analysis_icons_no_icons} shows results for how often a puzzle containing a specific rule is solved correctly by the best-performing model (GPT-4o). On \textsc{COLUMBUS-text}, GPT-4o performs the best on the \textit{relational} rules and the worst on \textit{individual} rules (difference of 17.98\%). When \textit{individual} rules appear together with \textit{modifier} rules, the model performance is slightly higher (by 3.02\%). We see a similar trend for \textsc{COLUMBUS-icon}, with a gap from \textit{individual} to \textit{relational} and \textit{modifier} rules being 10.96\% and 8.21\%, respectively. We note that, while the GPT-4o's performance is similar on the two partitions, specific rules are more difficult for this model when represented as text (e.g., \textit{repetition} rules). In contrast, others are more challenging when presented as icons (e.g., \textit{size}). Such biases align with recent work that shows the perceptual sensitivity of VLMs to object visual attributes~\cite{zhang2024exploring}. As for \textit{relational} rules, the model performance on text and icon puzzles is on par.

\subsection{VLM Generation of Puzzles}
\label{section_rebus_generation_results}

\begin{figure}[!ht]
    \centering
    \includegraphics[scale=0.59]{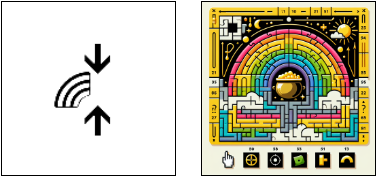}
    \caption{Rebus images for “\textit{end of the rainbow}”, generated by our automated pipeline (left) and by DALLE-3 (right).}
    \label{fig:generation}
\end{figure}

Given the strong generative abilities of VLMs, a natural question arises: can they generate puzzles without the methodology we define in Section \ref{section_methodology}? To investigate whether the taxonomy-based generation pipeline is necessary, we sample 100 puzzle answers and use DALLE-3~\cite{betker2023improving} to generate corresponding puzzles with the prompt ``Try to generate an image for a rebus puzzle on \{answer\}''. Three human annotators are asked first to label whether the puzzles contain sufficient visible elements to support solving the puzzle and then select the better one between the two puzzles (the one generated by our method and the one by DALLE-3). Our results show that 98\% of the rebuses generated by our pipeline contain a sound and complete list of elements, compared to only 44\% for DALLE-3. Additionally, the puzzles from our pipeline were preferred over those from DALLE-3 84\% of the time, with an 11\% tie rate. DALLE-3 struggles as a rebus generator for two main reasons~\Cref{fig:generation}~\cite{betker2023improving}: 1) Noisy details: unlike the concise rebuses our pipeline produces, DALLE-3 often creates very complex puzzles that include irrelevant information (e.g., the click icon in the right of \Cref{fig:generation}). 2) Abstract representation: DALLE-3 struggles to represent abstract ideas, such as ``after'', whereas our carefully designed rule-based taxonomy can handle these concepts precisely.

\section{Conclusions}

This paper defined a novel visual lateral thinking task associated with a construction methodology and a resulting synthetic, multiple-choice benchmark COLUMBUS consisting of 1005 rebus puzzles with text and icons. Our experiments on COLUMBUS indicated a substantial gap between human and model performance, which was lessened by adding the graph description to the model inputs. The improvement when expanding a model's access to include \textit{relational} rules implies they rely primarily on the elements in a puzzle alone rather than the spatial relationships between them. The analysis showed that models struggle to think laterally about rules that rearrange text because these require flexible, puzzle-specific abstractions.

The scale and difficulty of \textsc{COLUMBUS} is still limited. Moreover, its relative coverage of icon and text puzzles and various rules is imbalanced. Future work should apply and extend the methodology provided in this paper to develop more extensive and balanced versions of \textsc{COLUMBUS}, possibly by including other multimodal formats beyond rebus puzzles and reducing the variability between puzzle categories.
This can also be improved by increasing the flexibility of the generation process by making graph structures more complex to adjust puzzle difficulty and varying templates across perceptual dimensions like color, positioning, and size. 
Special attention should also be devoted to puzzles that require abstraction and creative thinking and cannot be solved easily by word matching a text/icon. While our methodology includes measures to address this (e.g., homophones and questions with multiple distractors containing the visible words of a puzzle), it can be extended by including synonyms or related concepts.










\clearpage
\clearpage
\bibliography{aaai25}

\clearpage
\appendix
\import{./}{appendix.tex}

\end{document}

%% file: appendix.tex
\begin{figure*}[!t]
    \centering
    \begin{subfigure}{0.25\textwidth}
      \centering
      \fbox{\includegraphics[width=0.7\linewidth]{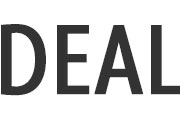}}
      \caption{Rebus puzzle that manipulates size (solution: \textit{Big Deal})}
      \label{fig:rebus_puzzle_big_deal}
    \end{subfigure}\hfil
    \begin{subfigure}{0.27\textwidth}
      \centering
      \fbox{\includegraphics[width=0.7\linewidth]{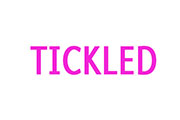}}
      \caption{Rebus puzzle that manipulates color (solution: \textit{Tickled Pink})}
      \label{fig:rebus_puzzle_tickled_pink}
    \end{subfigure}\hfil
    \begin{subfigure}{0.27\textwidth}
      \centering
      \fbox{\includegraphics[width=0.7\linewidth]{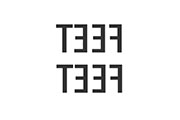}}
      \caption{Rebus puzzle that manipulates numbers (solution: \textit{Two Left Feet})}
      \label{fig:rebus_puzzle_two_left_feet}
    \end{subfigure}
    \\
    \begin{subfigure}{0.27\textwidth}
      \centering
      \includegraphics[width=0.7\linewidth]{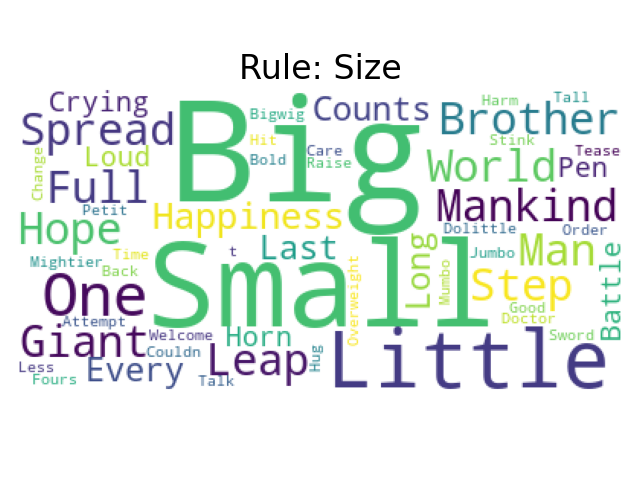}
      \caption{Word cloud for rebus puzzle answers that encode a rule for size.}
      \label{fig:rebus_type_wordcloud_size}
    \end{subfigure}\hfil
    \begin{subfigure}{0.27\textwidth}
      \centering
      \includegraphics[width=0.7\linewidth]{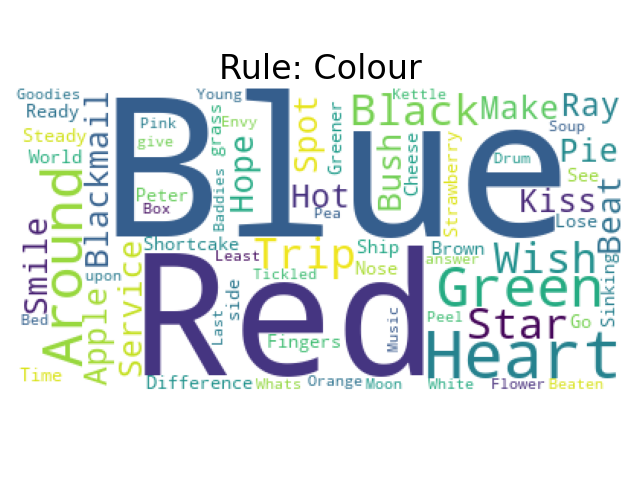}
        \caption{Word cloud for rebus puzzle answers that encode a color rule.}  
        \label{fig:rebus_type_wordcloud_colour}
    \end{subfigure}\hfil
    \begin{subfigure}{0.27\textwidth}
      \centering
      \includegraphics[width=0.7\linewidth]{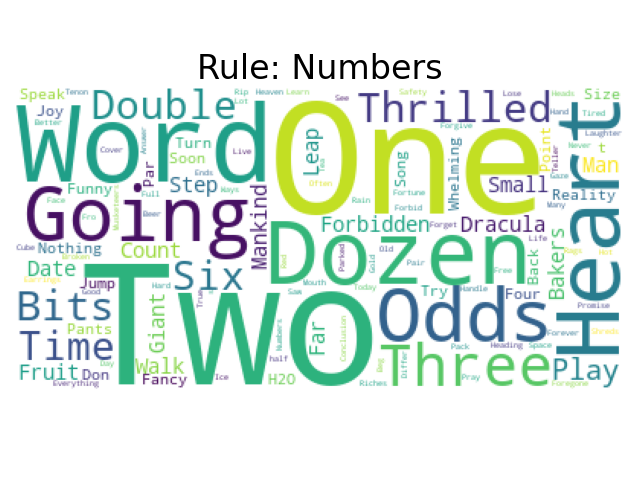}
        \caption{Word cloud for rebus puzzle answers that encode a rule for numbers.}  
        \label{fig:rebus_type_wordcloud_numbers}
    \end{subfigure}
    \caption{Examples showing how the words in a rebus puzzle answer relate to the puzzle itself. For each column, the top half image shows an example of a rebus puzzle that encodes the rule from the bottom half. The bottom half of the image is a word cloud consisting of all rebus puzzle answers that encode the specified rule. These puzzles and their answers were all scraped from an online database of puzzles (rebuses.co), each with a tag showing which rule is encoded.}
    \label{fig:examples_rebus_rules}
\end{figure*}

\section{Puzzle Generation}

\subsection{Data Filtering}
\label{data_filtering}

Of 18,836 puzzles, 4386 remain after automatic filtering (see Section  \ref{section_columbus_benchmark}). From this subset, we take the top 1000 puzzles according to those that have the most rules per node, plus the most edges (excluding the relational \textit{next to} rule). We expect these puzzles to be more challenging, as they require one to decode more about an element to understand its use in a puzzle. We manually remove unreadable puzzles that were broken from the generation process, such as too-large elements, overlapping text/icons, and overflowing text/icons. Then, we automatically remove puzzles in which the only rule used is replacing text with icons. As these puzzles are more similar to emoji puzzles than a rebus, we do not consider them to require a sufficient degree of lateral thinking. 


We perform minimal data filtering on the generated distractors. For the set of distractors belonging to each question, we only replace a distractor if it features another distractor dissimilar by at least one stopword that does not trigger any rules. In such a case, we replace it with the next most similar distractor (if it is also not too similar for the same reason). For example, the idioms ``Fresh start'' and ``A fresh start'' differ by the stopword ``A'', which does not trigger any rules, so one of these idioms can be replaced. If two words are the singular and plural forms of each other, we do not replace either, as plurality is used to trigger the \textit{repetition} rules. 

\subsection{Impact of Keywords} \label{section_appendix_keywords_impact}
Figure \ref{fig:examples_rebus_rules} shows three examples of rebus puzzles that highlight the relationship between the words in the answer of a rebus puzzle and what kind of rules are being triggered in the image of that rebus puzzle. For example, the rebus puzzle for the phrase \textit{Big Deal} (Figure \ref{fig:rebus_puzzle_big_deal}), focuses on size as a rule, which is triggered by the word \textit{Big}. Other keywords that tend to be involved in a rebus puzzle that focuses on size are shown in Figure \ref{fig:rebus_type_wordcloud_size}.

\subsection{Ignored Words} \label{appendix_ignored_words}
The list of ignored words is: ``the", ``a", ``of", ``is", ``let", ``and", ``at".

\subsection{Rule Keywords} \label{appendix_rule_keywords}
Tables \ref{tab:individual_rule_keywords} and \ref{tab:relational_rule_keywords} show the keywords that trigger the rules belonging to the \textit{individual} and \textit{relational} categories. 


\begin{table*}[!ht]
    \small
    \centering
    \caption{Keywords that trigger each \textit{individual} rule.}
    \begin{tabular}{@{}llllllllll@{}}
        \toprule
        \multicolumn{3}{c@{}}{{Direction}} & \multicolumn{3}{c@{}}{{Highlight}} & \multicolumn{2}{c@{}}{{Size}} & & \\
        \cmidrule(lr){1-3} \cmidrule(lr){4-6} \cmidrule(lr){7-8}
        {Up} & {Down} & {Reverse} & {Before} & {Middle} & {After} & {Big} & {Small} & {Color} & {Cross} \\
        \midrule
         
        Up & Down & Back & Before & Middle & After & Big & Little & Black & Cross \\
         &  & Mirror & Begin & Mid & End & Large & Micro & Blue & Crossed \\
         &  & Inverse & Start &  & Behind & Grand & Smaller & Orange & Crossing \\
         &  & Rear & Left &  &  & Bigger & Smallest & Green &  \\
         &  & Left & Starting &  &  & Biggest & Miniature & Red &  \\
         &  & Flip & Beginning &  &  & Giant &  & Purple &  \\
         &  &  &  &  &  & Jumbo &  & Brown &  \\
         &  &  &  &  &  &  &  & Pink &  \\
         &  &  &  &  &  &  &  & Gray &  \\
         &  &  &  &  &  &  &  & Yellow &  \\
         &  &  &  &  &  &  &  & Gold &  \\
        \bottomrule
    \end{tabular}
    \label{tab:individual_rule_keywords}
\end{table*}

\begin{table*}[!ht]
    \small
    \centering
    \caption{Keywords that trigger each \textit{relational} rule.}
    \begin{tabular}{@{}llllllllll@{}}
        \toprule
        {Next} to & {Inside} & {Above} & {Outside} \\
        \midrule
        Next & In     & Above & Out     \\
             & Inside & Over  & Outside \\
             & Into   & On    &         \\
             &        & Upon  &         \\
        \bottomrule
    \end{tabular}
    \label{tab:relational_rule_keywords}
\end{table*}




\subsection{Rule Frequency} \label{section_appendix_benchmark_overview}
Table \ref{tab:appendix_rule_sample_size} shows the frequency of each rule across all three categories, on both \textsc{COLUMBUS-text} and \textsc{COLUMBUS-icon}. 

\begin{table*}[!ht]
    \small
    \centering
    \caption{Table showing the sample size of puzzles that contain a specified rule for \textsc{COLUMBUS-text} and \textsc{COLUMBUS-icon}. The \textit{direction} rules are omitted for \textsc{COLUMBUS-icon} because these either do not function with icons, or become functionally identical to other rules when combined with icons.}
    \begin{tabular}{@{}llll@{}}
        \toprule
        \multirow{2}{*}{{Rule Type}} & \multirow{2}{*}{{Rule}} & \multicolumn{2}{c@{}}{\makecell[c]{{Rule sample size}}} \\
        \cmidrule(lr){3-4}
        & & {\textsc{text}} & {\textsc{icon}} \\
        \midrule
        \multirow{10}{*}{Individual} 
        & Direction: Up & 32 & N/A \\
        & Direction: Down & 26 & N/A \\
        & Direction: Reverse & 30 & N/A \\
        \cmidrule(lr){2-4}
        & Highlight: Before & 23 & 10 \\
        & Highlight: Middle & 19 & 10 \\
        & Highlight: After & 26 & 10 \\
        \cmidrule(lr){2-4}
        & Size: Big & 18 & 11 \\
        & Size: Small & 23 & 10 \\ 
        \cmidrule(lr){2-4}
        & Color & 27 & 13 \\
        \cmidrule(lr){2-4}
        & Cross & 30 & 12 \\
        \midrule
        \multirow{4}{*}{Relational} 
        & Next to & 293 & 227 \\
        \cmidrule(lr){2-4}
        & Inside & 122 & 104 \\
        \cmidrule(lr){2-4}
        & Above & 97 & 76 \\
        \cmidrule(lr){2-4}
        & Outside & 29 & 18 \\
        \midrule
        \multirow{3}{*}{Modifier} 
        & Sound & 288 & 150 \\
        \cmidrule(lr){2-4}
        & Repetition: Two  & 153 & 77 \\
        & Repetition: Four & 62 & 28 \\
        \bottomrule
    \end{tabular}
    \label{tab:appendix_rule_sample_size}
\end{table*}

\section{Implementation and Computation Details} \label{section_appendix_implementation_computation}
For image generation, each rebus puzzle is a 400 $\times$ 400 pixel image created in Matplotlib. We use Consolas's monospaced font for text, as it is convenient for spatial calculations. Icons are rendered in the Segoe UI Emoji font. 

For distractor generation, we use $\lambda = 0.8$ to prioritize word overlap similarity over semantic similarity in the weighted average calculation. The Sentence-BERT model used to calculate semantic similarity is all-MiniLM-L6-v2, adapted from \cite{wang_2020_minilm}. This is the most downloaded sentence similarity model available on Huggingface.

During inference, we do not alter the hyperparameters provided in the documentation of their Huggingface pages outside of the generated output token length where necessary. The models with altered output token lengths are Fuyu-8b (200 tokens), Mistral (100 tokens), and the two Llava models (200 tokens).
All models were evaluated on an in-house Rocky Linux operating system cluster. 
Almost all models were run on a Nvidia RTX A6000 (48 GB), except BLIP-2 OPT-2.6b, which was used on an RTX A10 (24 GB). For the closed-source models, we use the APIs provided by OpenAI and Google, which were run on a Linux machine with default hyperparameter settings.

Almost all randomness prevalent in the experiments uses a seed of 42. For backward chaining, these results are calculated using a seed of 43. We computed the backward chaining results on a random subset of 50 puzzles because answering each puzzle takes approximately 20 API calls, which incurs prohibitive time and money costs. 

We use almost all the same hyperparameters outlined by \citeauthor{kassner_2023} for experiments with backward chaining through belief graphs. However, the maximum depth we set is 1 due to the relatively simple structure of rebus puzzles in COLUMBUS. Additionally, $c_{\mbox{xor}}$ and $c_{\mbox{mc}}$ (the weights for the XOR and multiple-choice rule nodes in a belief graph) are both set to 1 based on manual tuning.

\section{Prompt Strategies} 

\subsection{Templates} \label{section_appendix_prompt_templates}
As described in the paper, we design four human-curated prompts in increasing order of information supplied to a model.
Table \ref{tab:prompt_templates} provides the exact prompt templates used. 

\begin{table*}[!ht]
    \small
    \centering
    \caption{Prompt templates used during experimentation, as seen in Tables \ref{tab:results_main_table} and \ref{tab:results_prompt_table}, and Figures \ref{fig:results_all_prompts_four_models} and \ref{fig:results_rule_analysis_icons_no_icons}. The sections highlighted in bold are the new information added to each prompt from the previous prompt.}
    \begin{tabular}{@{}lp{14.5cm}@{}}
        \toprule
        {Prompt} & {Template} \\
        \midrule
        \makecell[l]{Prompt 1\\(no knowledge)} & \textit{[IMG]} Which word/phrase is conveyed in this image from the following options (either A, B, C, or D)?\newline (A) \{\} (B) \{\} (C) \{\} (D) \{\} \\
        \midrule
        \makecell[l]{Prompt 2\\(puzzle\\knowledge)} & \textit{[IMG]} \textbf{You are given a rebus puzzle. It consists of text or icons that is used to convey a word or phrase. It needs to be solved through creative thinking.} Which word/phrase is conveyed in this image from the following options (either A, B, C, or D)?\newline(A) \{\} (B) \{\} (C) \{\} (D) \{\} \\
        \midrule
        \makecell[l]{Prompt 3\\(graph node\\knowledge)} & \textit{[IMG]} You are given an image of a rebus puzzle. It consists of text or icons that is used to convey a word or phrase. It needs to be solved through creative thinking. \textbf{You are also given a description of the graph representation of the puzzle. The nodes are elements that contain text or icons, which are then manipulated through the attributes of their node. The description is as follows:\newline \{\}}\newline Which word/phrase is conveyed in this image and description from the following options (either A, B, C, or D)?\newline(A) \{\} (B) \{\} (C) \{\} (D) \{\} \\
        \midrule
        \makecell[l]{Prompt 4\\(graph nodes + \\edges knowledge)} & \textit{[IMG]} You are given an image of a rebus puzzle. It consists of text or icons that is used to convey a word or phrase. It needs to be solved through creative thinking. You are also given a description of the graph representation of the puzzle. The nodes are elements that contain text or icons, which are then manipulated through the attributes of their node. \textbf{The edges define spatial relationships between these elements.} The description is as follows:\newline \{\}\newline Which word/phrase is conveyed in this image and description from the following options (either A, B, C, or D)?\newline(A) \{\} (B) \{\} (C) \{\} (D) \{\} \\
        \bottomrule
    \end{tabular}
    \label{tab:prompt_templates}
\end{table*}

\subsection{Graph Descriptions}
Table \ref{tab:prompt_graph_descriptions} shows two descriptions for the graphs in Figure \ref{fig:rebus_graph_examples}, fed to a model through prompts 3 and 4.

\begin{table*}[!ht]
    \small
    \centering
    \caption{Descriptions of graphs used in prompts 3 and 4 from Table \ref{tab:prompt_templates}. For clarity, the two examples from Figure \ref{fig:rebus_graph_examples} are used in this table.}
    \begin{tabular}{@{}p{1.74cm}p{7.5cm}p{7.5cm}@{}}
        \toprule
        {Input} & {Graph description (only nodes)} & {Graph description (nodes + edges)} \\
        \midrule
        
        

        
        \textit{Blueprints} & Node 1 attributes: (text: PRINT, color: blue, repeat: 2) & Node 1 attributes: (text: PRINT, color: blue, repeat: 2)
        \\
        \midrule
        \textit{Pull the wool over eyes}
        &
        Node 1 attributes: (text: PULL, repeat: 1)
        
        Node 2 attributes: (text: WOOL, repeat: 1)
        
        Node 3 attributes: (text: IIII, repeat: 1, sound: (eyes: iiii))
        &
        Node 1 attributes: (text: PULL, repeat: 1)
        
        Node 2 attributes: (text: WOOL, repeat: 1)
        
        Node 3 attributes: (text: IIII, repeat: 1, sound: (eyes: iiii))

        Edge 1: node 1 to node 2 (rule: NEXT-TO)
        
        Edge 2: node 2 to node 3 (rule: ABOVE)
        \\
        \bottomrule
    \end{tabular}
    \label{tab:prompt_graph_descriptions}
\end{table*}

\section{Human Evaluation} \label{section_appendix_human_evaluation}
\begin{figure*}
    \centering
    \includegraphics[scale=0.5]{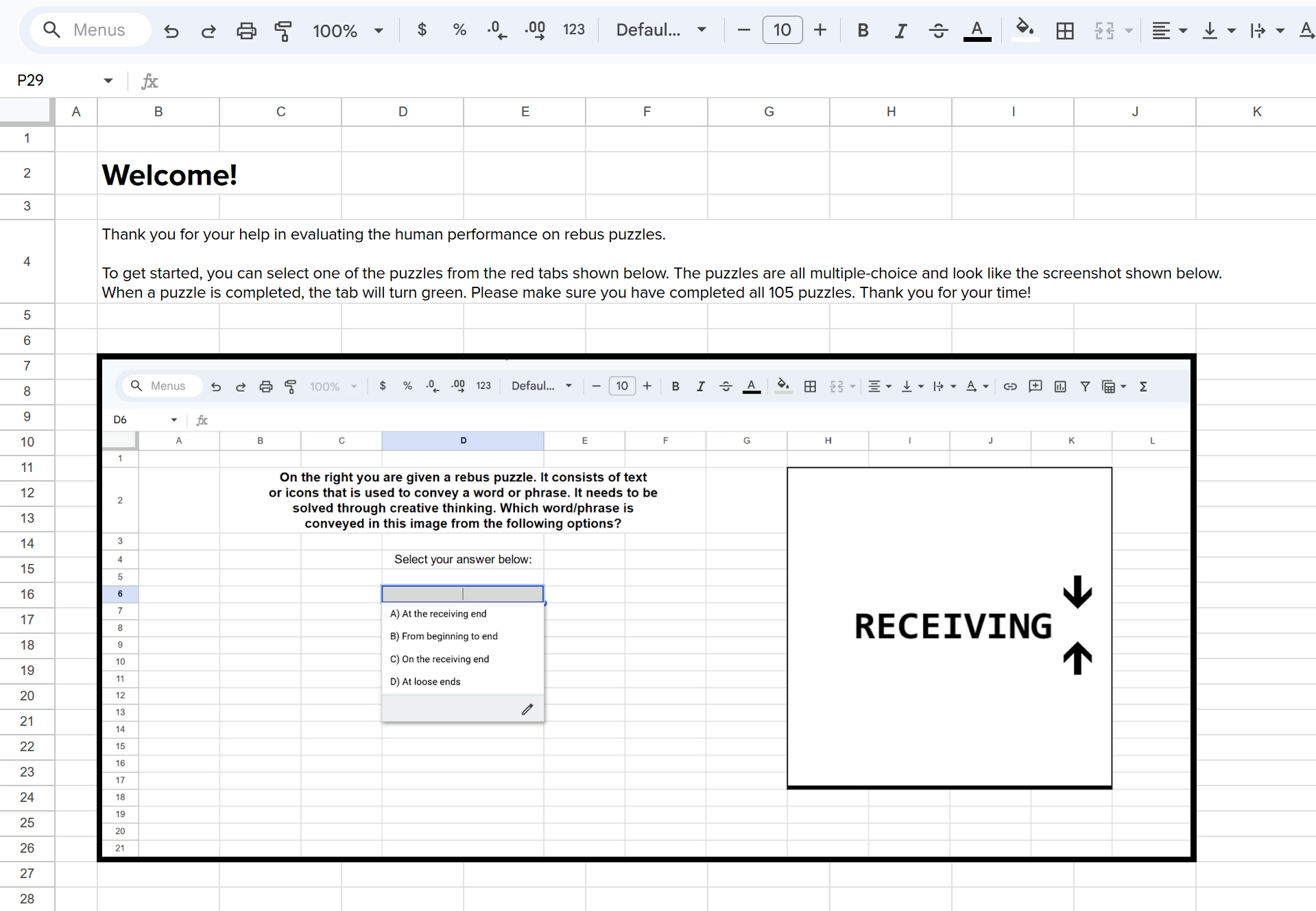}
    \caption{Page displayed before starting.}
    \label{fig:appendix_launch_page}
\end{figure*}

\begin{figure*}
    \centering
    \includegraphics[scale=0.4]{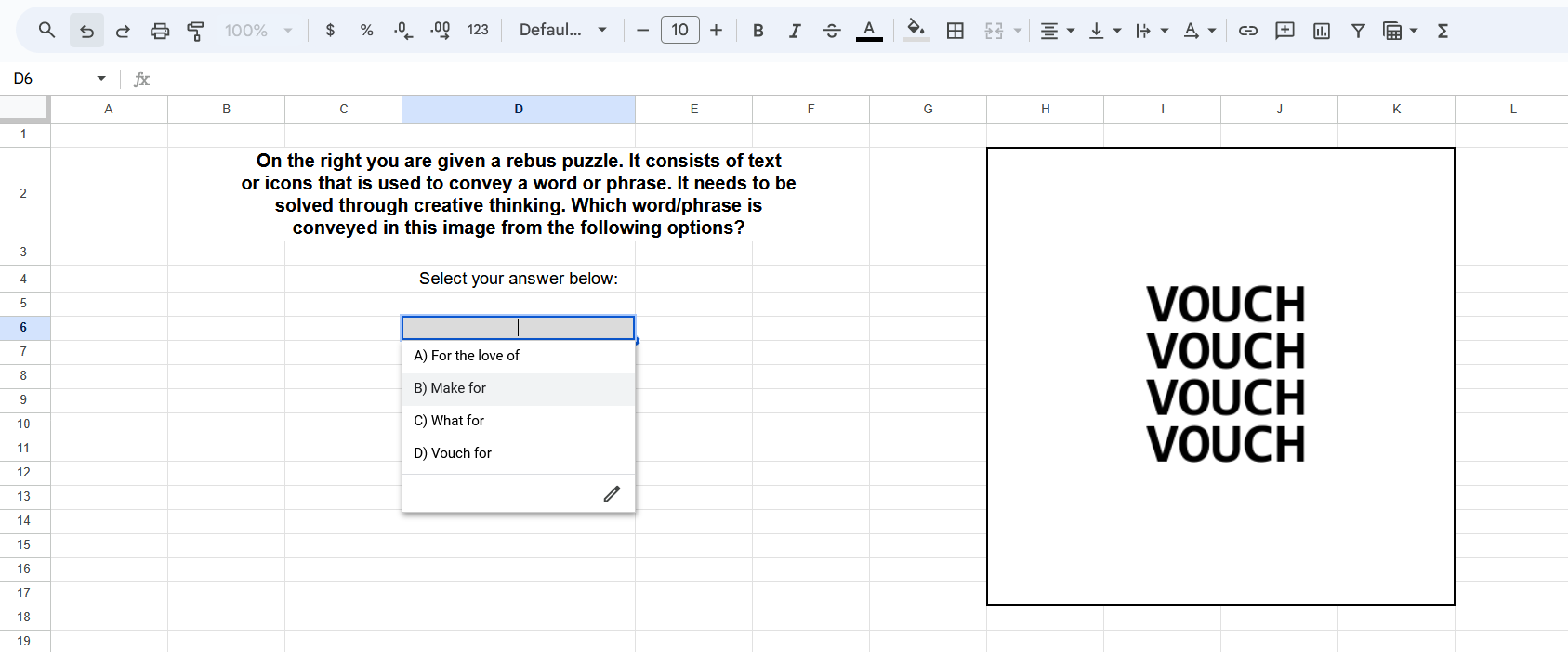}
    \caption{Page displayed for each puzzle.}
    \label{fig:appendix_puzzle_page}
\end{figure*}

Figures \ref{fig:appendix_launch_page} and \ref{fig:appendix_puzzle_page} show the launch and puzzle pages, respectively, that a human participant sees while answering puzzles in the benchmark. This setup is implemented with Google Sheets.

\section{Additional Results} \label{section_additional_results}
\subsection{Model Results per Prompt}
Table \ref{tab:results_prompt_table} shows the extended results across all models for each of the four prompts.

\begin{table*}[h]
    \small
    \centering
    \caption{Table showing the mean accuracy (\%) averaged across three runs for four prompts that supply the model with varying levels of information (see Appendix \ref{section_appendix_prompt_templates}). The results from Table \ref{tab:results_main_table} are also shown with prompt 2. Results for Mistral are also reported, but only including the graph description (no image is passed). Highest results in each column are in bold. The models are grouped according to their category outlined in Section \ref{section_model_selection} (open-source non-instruction VQA, open-source instruction VQA, closed-source VQA, instruction QA).}
    \begin{tabular}{@{}lllllllll@{}}
        \toprule
        \multirow{2}{*}{Model / Prompt} 
        & \multicolumn{4}{c@{}}{
        \makecell[c]{COLUMBUS-\textsc{text}}}
        & \multicolumn{4}{c@{}}{\makecell[c]{COLUMBUS-\textsc{icon}}} \\
        \cmidrule(lr){2-5} \cmidrule(lr){6-9}
        & 1 & 2 & 3 & 4 & 1 & 2 & 3 & 4 \\
        \midrule
        
        \makecell[l]{BLIP-2 OPT (2.7b)} 
        & 24.7 & 21.9 & 18.9 & 21.5 & 23.2 & 24.1 & 16.6 & 21.1 \\
        
        \makecell[l]{BLIP-2 OPT (6.7b)} 
        & 22.2 & 24.0 & 23.2 & 23.6 & 25.9 & 25.6 & 22.6 & 20.8 \\
        
        Fuyu (8b) 
        & 32.4 & 32.0 & 42.9 & 45.5 & 32.6 & 31.0 & 47.4 & 49.3 \\
        \midrule
        
        \makecell[l]{InstructBLIP Vicuna (7b)}
        & 50.8 & 51.5 & 45.3 & 43.3 & 51.9 & 51.8 & 39.6 & 38.6 \\
        
        Qwen-VL (7b)
        & 56.7 & 57.8 & 67.3 & 56.0 & 67.1 & 63.2 & 72.0 & 64.6 \\
        
        \makecell[l]{BLIP-2 Flan-T5-XXL (11b)} 
        & 64.8 & 68.2 & 85.1 & 84.9 & 67.1 & 72.0 & 91.6 & 90.8 \\
        
        Llava (13b)
        & 63.2 & 58.0 & 70.9 & 71.1 & 61.2 & 58.8 & 69.5 & 69.1 \\
        
        CogVLM (17b)
        & 59.9 & 59.3 & 63.5 & 64.0 & 59.3 & 60.1 & 63.9 & 64.7 \\
        
        Llava (34b) 
        & 67.9 & 66.8 & 78.7 & 79.2 & 71.3 & 73.1 & 85.5 & 85.4 \\
        \midrule

        GPT-4o & \textbf{75.8} & \textbf{80.9} & \textbf{90.6} & \textbf{90.9} & \textbf{76.4} & \textbf{83.3} & \textbf{93.6} & \textbf{93.2} \\

        GPT-4o-mini & 67.8 & 74.0 & 87.0 & 86.3 & 66.4 & 77.7 & 88.3 & 88.9 \\

        \makecell[l]{Gemini 1.5 (Pro)} & 68.9 & 71.6 & 87.5 & 88.4 & 72.9 & 77.5 & 90.6 & 91.8 \\

        \makecell[l]{Gemini 1.5 (Flash)} & 63.6 & 64.4 & 86.4 & 87.0 & 67.4 & 67.4 & 86.7 & 87.1 \\

        \midrule
    
        Mistral (7b) & N/A & N/A & 70.6 & 75.3 & N/A & N/A & 73.7 & 76.8 \\
        \bottomrule
    \end{tabular}
    \label{tab:results_prompt_table}
\end{table*}

\section{Code Appendix}
The code for generating a rebus graph from a compound word or phrase (see Section \ref{section_puzzle_generation}) is shown in Listings \ref{lst:compound_parser_parse} and \ref{lst:phrase_parser_parse}, respectively.

\begin{listing*}[tb]%
\caption{Code to generate a rebus graph for a compound word.}%
\label{lst:compound_parser_parse}%
\begin{lstlisting}[language=Python,escapeinside=``]
def parse(self, c1, c2, is_plural):
    """
    Generates all possible single-node graphs from the specified input words (c1, c2).

    :param c1: first word that will be checked to see if it triggers any rule keywords.
    :param c2: second word that will be checked to see if it triggers any rule keywords.
    :param is_plural: flag to denote if combined word (i.e., f"{c1}{c2}") is plural. This is used to trigger the repetition rules. 
    NOTE: this only applies to compound words, not for pairs of words.
    :return: a list of single-node graphs corresponding to each of the generated graphs.
    """
    
    # Check for patterns for either constituent word
    rules_c1 = Rule.find_all(c1, is_plural)
    rules_c2 = Rule.find_all(c2, is_plural)

    # List to hold all possible generated puzzles
    graphs = []

    # Generate puzzles by combining both words into one
    for rule_c1 in rules_c1:
        graphs += self._generate_rebus(c2, rule_c1, is_plural)
    for rule_c2 in rules_c2:
        graphs += self._generate_rebus(c1, rule_c2, is_plural)

    # Generate puzzles by placing both words next to each other
    graphs += self._generate_rebus(c1, {}, is_plural, c2)

    # Remove duplicate puzzles
    graphs = remove_duplicate_graphs(graphs)
    
    return graphs
\end{lstlisting}
\end{listing*}
\clearpage
\begin{listing*}[tb]%
\caption{Code to generate a rebus graph for a phrase.}%
\label{lst:phrase_parser_parse}%
\begin{lstlisting}[language=Python,escapeinside=``]
def parse(self, phrase):
    """
    Parses a phrase to its rebus graph representation.
    
    :param phrase: phrase to convert to a rebus graph.
    :return: list containing all the possible combinations of rebus graphs for the specified phrase.
    """
    
    # Remove ignored words from the phrase
    phrase_words = [word for word in phrase.split() if word not in self._ignore_words]
    phrase = " ".join(phrase_words)
    
    # For each word (pair) in the phrase, generate the possible graphs for that word (pair).
    graphs_per_word = self._get_all_graphs_per_word(phrase)
    all_graphs = self._get_all_combinations(graphs_per_word)
    
    return all_graphs
\end{lstlisting}
\end{listing*}
